\documentclass[a4paper]{article}
\usepackage{pstricks,pst-node}
\usepackage{amssymb,calbf,types,sorts2e,ded}
\usepackage{ijcai97}

\def\eqs{=^s}\def\eqc{=^{p}}
\def\typebool{t}\def\typeind{e}

\def\explicitsort#1{{\mbox{\sc{#1}}}}
\def\sortentity{\explicitsort{Entity}}
\def\sortabstract{\explicitsort{Abstract}}
\def\sortreal{\explicitsort{Real}}
\def\sortnumber{\explicitsort{Number}}
\def\sortinanimate{\explicitsort{Inanimate}}
\def\sortpen{\explicitsort{Pen}}
\def\sortgame{\explicitsort{Game}}
\def\sortanimate{\explicitsort{Animate}}
\def\sortanimal{\explicitsort{Animal}}
\def\sortdog{\explicitsort{Dog}}
\def\sorthuman{\explicitsort{Human}}
\def\sortman{\explicitsort{Man}}
\def\sortwoman{\explicitsort{Woman}}
\def\sortmale{\explicitsort{Male}}
\def\sortfemale{\explicitsort{Female}}
\def\sortsocial{\explicitsort{Social}}
\def\sortfriendly{\explicitsort{Friendly}}
\def\sortemotional{\explicitsort{Emotional}}

\def\cola{{\sf a}}
\def\colb{{\sf b}}
\def\colc{{\sf c}}
\def\cold{{\sf d}}
\def\cole{{\sf e}}

\def\colpe{{\sf pe}}

\def\colnpe{{\neg\sf pe}}

\def\colavar{{\sf A}}
\def\colbvar{{\sf B}}

\newcommand{\bc}{\begin{center}}
\newcommand {\ec}{ \end{center}}
\newcommand {\be} {\begin{enumerate}}
\newcommand {\ee} {\end{enumerate}}
\newcommand {\bi} {\begin{itemize}}
\newcommand {\ei} {\end{itemize}}
\newcommand {\ba} {\vspace*{3pt}\noindent$\begin{array}}
\newcommand {\et} {\end{tabular}\vspace*{3pt}}
\newcommand {\bt} {\vspace*{3pt}\noindent\begin{tabular}}
\newcommand {\ea} {\end{array}$\vspace*{3pt}}
\newcommand {\txt}[1]{\mbox{{\it{#1}}}}
\def\wff{\txt{wff}}

\def\Colours{{\cal C}}
\def\Covars{{\cal C\kern-.2ex V}}
\def\Col{\mbox{$\cal C$}}

\def\lam{\lambda}
\def\ov#1{\overline{#1}}

\title{Computing Parallelism in Discourse}
\author{Claire Gardent \\
Computational Linguistics \\ Universit\"at des Saarlandes\\
Saarbr\"ucken, Germany \\
{\tt claire@coli.uni-sb.de}\And
Michael Kohlhase \\
Computer
Science \\ Universit\"at des Saarlandes \\
Saarbr\"ucken, Germany \\
{\tt kohlhase@cs.uni-sb.de}}

\begin{document}
\maketitle
\begin{abstract}
  Although much has been said about parallelism in discourse, a
  formal, computational theory of parallelism structure is still
  outstanding. In this paper, we present a theory which given two
  parallel utterances predicts which are the parallel elements. The
  theory consists of a sorted, higher-order abductive calculus and we
  show that it reconciles the insights of discourse theories of
  parallelism with those of Higher-Order Unification approaches to
  discourse semantics, thereby providing a natural framework in which
  to capture the effect of parallelism on discourse semantics.
\end{abstract}

\section{Introduction}
\label{sec:intro}
Both Higher-Order Unification (HOU) approaches to discourse semantics
\cite{DalShiPer:eahou91,ShiPerDal:iosae96,GaKoLe:cahou96,GaKo:hocuanls96}
and discourse theories of parallelism
\cite{Hobbs:lac90,Kehler:icfitcodi95} assume parallelism
structuration: given a pair of parallel utterances, the parallel
elements are taken as given.

This assumption clearly undermines the predictive power of a
theory. If parallel elements are stipulated rather than predicted,
conclusions based on parallelism remain controversial: what would
happen if the parallel elements were others? And more crucially, what
constraints can we put on parallelism determination (when can we rule
out a pair as not parallel?)

In this paper, we present a theory of parallelism which goes one step
towards answering this objection. Given two utterances, the theory
predicts which of the elements occurring in these utterances are
parallel to each other. The proposed theory has one additional
important advantage: it incorporates HOU as a main component of
parallelism theory thereby permitting an integration of the HOU
approach to discourse semantics with discourse theories of
parallelism. The resulting framework permits a natural modelling of
the often observed effect of parallelism on discourse semantics
\cite{Lang:sdkv77,Asher:rtaoid93}. We show in particular that it
correctly captures the interaction of VP-ellipsis and gapping with
parallelism. More generally however, the hope is that it also provides
an adequate basis for capturing the interaction of parallelism with
such discourse phenomena as deaccenting, anaphora and quantification.

Our approach departs from~\cite{GroBreManMoe:puagidg94} in
that it genuinely can predict parallel elements. Although both
approaches rely on a sorts/types hierarchy and on some kind of
overwriting to carry out this task, our approach readily extends to
gapping cases whereas as we shall see (cf.
section~\ref{sec:gapping})~\cite{GroBreManMoe:puagidg94} doesn't.

We proceed as follows. First we present a sort-based abductive
calculus for parallelism and show that it predicts parallel elements.
We then show how this abductive calculus can be combined with HOU
thus yielding an integrated treatment of parallelism and discourse
semantics. We then conclude with pointers to further research and
related work.

\section{Defining discourse parallelism} 
\label{s2}

In linguistic theories on discourse
coherence~\cite{Kehler:icfitcodi95}, ellipsis~\cite{DalShiPer:eahou91}
(henceforth DSP) and corrections~\cite{GaKoLe:cahou96}, the notion of
parallelism plays a central role. In particular, the HOU-based
approaches presuppose a theory of parallelism which precomputes the
parallel elements of a pair of utterances. For instance, given the
utterance pair {\it Jon likes golf. Peter does too}, DSP's analysis of
ellipsis presupposes that {\it Jon} and {\it Peter} have been
recognised as being parallel to each other.

Similarly, discourse theories of parallelism also assume parallelism
structuration. According to \cite{Hobbs:lac90,Kehler:icfitcodi95} for
instance, there is a class of discourse relations (the {\it
  resemblance relations}) which involve the inferring of structurally
parallel propositions and where arguments and predicates stand in one
of the following configurations:
\begin{center}
\begin{tabular}{llll}
Relation & S-Ent & T-Ent & Reqts \\
Parallel & $p(\vec{a})$ &  $p(\vec{b})$ & $a_i, b_i$ similar \\
Contrast & $p(\vec{a})$ &  $\neg p(\vec{b})$ & $a_i, b_i$ similar \\
 & $p(\vec{a})$ &  $p(\vec{b})$ & $a_i, b_i$ contrastive \\
Exemplification & $p(\vec{a})$ &  $p(\vec{b})$ & $a_i\in b_{1}$ or $a_{i}\subseteq b_i$ \\
Generalisation & $p(\vec{a})$ &  $p(\vec{b})$ &  $b_i\in a_{1}$ or $b_{i}\subseteq a_i$
\end{tabular}
\end{center}
where $\vec{a}, \vec{b}$ represent argument sequences; $a_i , b_i$ are
any elements of these sequences; and S- and T-Ent are the propositions
entailed by the two (source and target) parallel utterances.
Furthermore, entities are taken to be {\it similar} if they share some
reasonably specific property and {\it contrastive} if they have both a
shared and a complementary property.

Again, the parallel elements ($a_i$ and $b_i$) are taken as given that
is, the way in which they are recognised is not specified. In what
follows, we present a computational theory of parallelism which
predicts these parallel elements.  The model is a simple abductive
calculus which captures Hobbs and Kehler's notions of parallelism and
constrast as they are given above. We make the simplifying assumptions
that contrast and parallelism are one and the same notion (we speak of
contrastive or {\it c-parallelism}) and that the properties $p$ used in
determining them are restricted to sorts from a given,
domain-specific sort hierarchy.  Thus we can use sorted type
theory~\cite{Kohlhase:amosho94} to model similarity and contrastive
parallelism.

\subsection{Sorted Logic}
\label{sec:sortlog}

Sorts correspond to the basic cognitive concepts. Logically they can either be seen
as unary predicates or as refinements of the types. The intuition behind this is that
the universe of objects of a type $\typea$ is subdivided in subsets which are
represented by sorts $\sorta,\sortb,\ldots$. Since these can in turn be subdivided
into subsets, the sorts are ordered by a partial ordering relation $\ior$ in a
so-called {\bf sort hierarchy}\footnote{For the purposes of this paper, we assume the
  sort hierarchy to be given. For applications, hierarchies could be generated from
  domain representations in KL-ONE like formalisms commonly used in
  NL systems.}

Just as in the case of types, every formula has a sort, that can be computed from the sorts of the
constants and variables occurring in it. In fact, formulae can have multiple sorts, corresponding to the
fact that the intersection of the sets represented by their sorts can be non-empty.

For this paper we assume a fixed finite set of sorts for each type. For
the base type $e$, we will use the following sort hierarchy in our examples. 

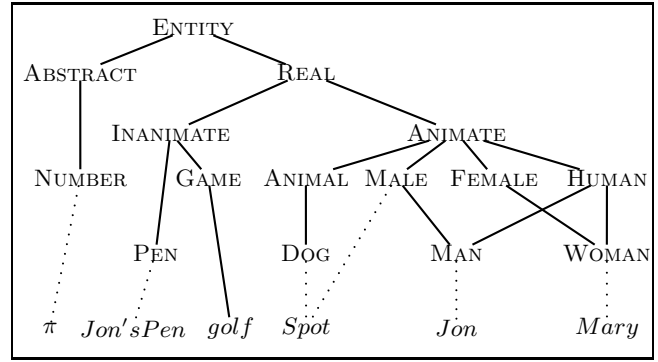
\begin{figure}[htbp]
\begin{center}
\leavevmode\footnotesize
\fbox{\begin{pspicture}(-0.8,-0.3)(7.5,4.2)
  \rput(-0.4,0){\rnode{pi}{$\pi$}}
  \rput(0,2){\rnode{num}{\sortnumber}}
  \rput(0,3.4){\rnode{abs}{\sortabstract}}
  \rput(0.7,0){\rnode{jpen}{$Jon's Pen$}}
  \rput(1,1){\rnode{pen}{\sortpen}}
  \rput(1.2,2.6){\rnode{inanimate}{\sortinanimate}}
  \rput(1.5,4){\rnode{ent}{\sortentity}}
  \rput(3,0){\rnode{spot}{$Spot$}}
  \rput(3,1){\rnode{dog}{\sortdog}}
  \rput(3,2){\rnode{animal}{\sortanimal}}
  \rput(1.7,2){\rnode{game}{\sortgame}}
  \rput(2,0){\rnode{golf}{$golf$}}
  \rput(4.2,2){\rnode{male}{\sortmale}}
  \rput(5.5,2){\rnode{female}{\sortfemale}}
  \rput(7,2){\rnode{human}{\sorthuman}}
  \rput(5,1){\rnode{man}{\sortman}}
  \rput(7,1){\rnode{woman}{\sortwoman}}
  \rput(5,2.6){\rnode{animate}{\sortanimate}}
  \rput(3,3.4){\rnode{real}{\sortreal}}
  \rput(5,0){\rnode{jon}{$Jon$}}
  \rput(7,0){\rnode{mary}{$Mary$}}
  \ncline[linestyle=dotted]{-}{pi}{num}
  \ncline[linestyle=dotted]{-}{jpen}{pen}
  \ncline[linestyle=dotted]{-}{spot}{dog}
  \ncline[linestyle=dotted]{-}{spot}{male}
  \ncline[linestyle=dotted]{-}{jon}{man}
  \ncline[linestyle=dotted]{-}{mary}{woman}
  \ncline{-}{num}{abs}
  \ncline{-}{abs}{ent}
  \ncline{-}{real}{ent}
  \ncline{-}{inanimate}{real}
  \ncline{-}{pen}{inanimate}
  \ncline{-}{game}{inanimate}
  \ncline{-}{game}{golf}
  \ncline{-}{dog}{animal}
  \ncline{-}{human}{man}
  \ncline{-}{male}{man}
  \ncline{-}{human}{woman}
  \ncline{-}{female}{woman}
  \ncline{-}{animal}{animate}
  \ncline{-}{human}{animate}
  \ncline{-}{male}{animate}
  \ncline{-}{female}{animate}
  \ncline{-}{animate}{real}
\end{pspicture}}
\caption{Sort hierarchy of type $\typeind$}
\label{fig:sort-ind}
\end{center}
\end{figure}

Note that the intersection of the sorts {\sortmale} and {\sortdog} is
non-empty, since the constant $Spot$ has both sorts. If we want to
make this explicit, we can give $Spot$ the {\bf intersection sort}
${\sortmale}\&{\sortdog}$. Even though we assume the {\bf simple
  sorts} (i.e. the non-intersection ones) to be non-empty, the
intersection sorts can in general be. For instance, the sorts
{\sortanimate} and {\sortinanimate} are disjoint, since they are
complementary.  The existence of complementary sorts allows us to
model the requirements for parallel elements quite naturally. Two
formulae $\bA$ and $\bB$ (of any type) are {\bf similar} if they have
a common sort; they are {\bf contrastive}, if they have a {\bf
  distinguishing sort} $\sortd$, i.e. if $\bA$ has sort $\sortd$ but
$\bB$ has sort $\neg\sortd$ or vice versa and finally they are {\bf
  c-parallel}, iff they are both.

For instance $Jon$ and $Mary$ are parallel, since both are of sort
{\sorthuman}, but $Jon$ has sort {\sortman}, whereas $Mary$ has sort
${\sortwoman}=\neg{\sortman}\&{\sorthuman}\&{\sortfemale}\ior\neg{\sortman}$
and therefore $Mary$ also has the distinguishing sort
$\neg{\sortman}$. This supports DSP's analysis of
\begin{quote}
  {\it Jon likes golf, and Mary likes golf.}
\end{quote}

For the higher-type, the sort hierarchies of lower type induce further
sorts: For any sorts $\sorta$ and $\sortb$ of types $\typea$ and
$\typeb$, $\sorta\ar\sortb$ is a {\bf functional} sort of type
$\typea\ar\typeb$. We call sorts that do not contain an arrow {\bf
  basic sorts}. Similarly, the sort hierarchy of lower type induces
subsort relations: $\sortb\ar\sortc\ior\sorta\ar\sortd$, is entailed
by $\sorta\ior\sortb$ and $\sortc\ior\sortd$.  Furthermore, the
resulting sorts can be further subdivided by {\bf functional base
  sorts}, i.e. sorts that do not contain an arrow, but are of
functional type.

For our examples we will use the following sort hierarchy of type $\typeind\ar\typeind\ar\typebool$
\begin{figure}[htbp]
  \begin{center}
    \leavevmode\small
\fbox{\begin{pspicture}(-1,-0.2)(7.5,2.2)
  \rput(0,0){\rnode{support}{$support$}} 
  \rput(2,0){\rnode{oppose}{$oppose$}}
  \rput(4,0){\rnode{like}{$like$}} 
  \rput(6,0){\rnode{dislike}{$dislike$}}
  \rput(0,1){\rnode{social}{\sortsocial}} 
  \rput(2,1){\rnode{friendly}{\sortfriendly}}
  \rput(4,1){\rnode{unfriendly}{$\neg\sortfriendly$}} 
  \rput(6.5,1){\rnode{emotional}{\sortemotional}}
  \rput(3.5,2){\rnode{hht}{$\sorthuman\ar\sorthuman\ar\typebool$}} 
  \ncline[linestyle=dotted]{-}{support}{social}
  \ncline[linestyle=dotted]{-}{support}{friendly}
  \ncline[linestyle=dotted]{-}{oppose}{social}
  \ncline[linestyle=dotted]{-}{oppose}{unfriendly}
  \ncline[linestyle=dotted]{-}{like}{friendly}
  \ncline[linestyle=dotted]{-}{like}{emotional}
  \ncline[linestyle=dotted]{-}{dislike}{unfriendly}
  \ncline[linestyle=dotted]{-}{dislike}{emotional}
  \ncline{-}{social}{hht}
  \ncline{-}{friendly}{hht}
  \ncline{-}{unfriendly}{hht}
  \ncline{-}{emotional}{hht}
\end{pspicture}}
    \caption{Sort hierarchy of type  $\typeind\ar\typeind\ar\typebool$}
    \label{fig:sort-eet}
  \end{center}
\end{figure}

\subsection{Computation of parallelism}
\label{sec:parcomp}

\begin{figure}[htbp]
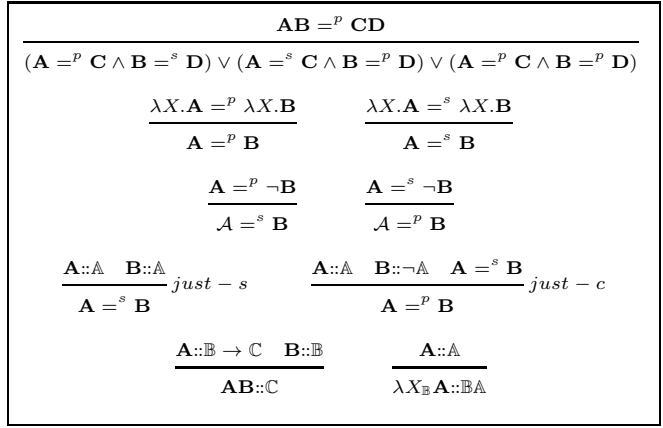

  \begin{center}
    \leavevmode\scriptsize
    \fbox{\hspace{1mm}\begin{textnd}
      \ianc{\bA\bB\eqc\bC\bD}
           {(\bA\eqc\bC\land\bB\eqs\bD)\lor
            (\bA\eqs\bC\land\bB\eqc\bD)\lor
            (\bA\eqc\bC\land\bB\eqc\bD)}{} \\
      \ianc{\lam X.\bA\eqc\lam X.\bB}
           {\bA\eqc\bB}{}\hspace{3em}
      \ianc{\lam X.\bA\eqs\lam X.\bB}
           {\bA\eqs\bB}{}                  \\
      \ianc{\bA\eqc\neg\bB}
           {\cA\eqs\bB}{} \hspace{3em}
      \ianc{\bA\eqs\neg\bB}
           {\cA\eqc\bB}{} \\
      \ibnc{\bA\ofsort\sorta}
           {\bB\ofsort\sorta}
           {\bA\eqs\bB}
           {just-s} \hspace{3em}
      \icnc{\bA\ofsort\sorta}
           {\bB\ofsort\neg\sorta}
           {\bA\eqs\bB}
           {\bA\eqc\bB}
           {just-c} \\
      \ibnc{\bA\ofsort\sortb\ar\sortc}
           {\bB\ofsort\sortb}
           {\bA\bB\ofsort\sortc}{} \hspace{3em}
      \ianc{\bA\ofsort\sorta}
           {\lambda X_\sortb\bA\ofsortb\sorta}{}
    \end{textnd}\hspace{1mm}}
    \caption{The abductive Calculus for Parallelism}
    \label{fig:parallel-calc}
  \end{center}
\end{figure}

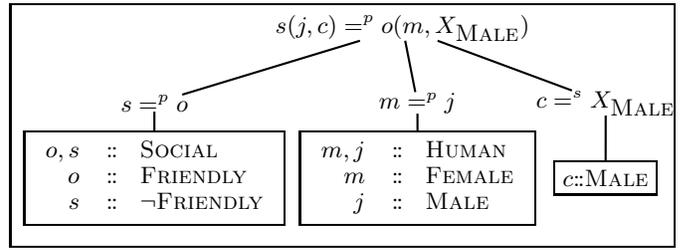
\begin{figure}[htbp]
  \begin{center}
    \leavevmode\footnotesize
    \fbox{\begin{pspicture}(-1.6,-0.8)(7,2.2)
      \rput(0.2,0){\rnode{sco-abd}{\fbox{$
          \begin{array}[t]{rcl}
            o,s&\ofsort&\sortsocial\\
            o&\ofsort&\sortfriendly\\
            s&\ofsort&\neg\sortfriendly
          \end{array}$}}}
      \rput(3.7,0){\rnode{jcm-abd}{\fbox{$
          \begin{array}[t]{rcl}
             m,j&\ofsort&\sorthuman\\
             m&\ofsort&\sortfemale\\
             j&\ofsort&\sortmale
           \end{array}$}}}
      \rput(6.2,0){\rnode{csx-abd}{\fbox{$c\ofsort\sortmale$}}}
      \rput(0.2,1){\rnode{sco}{$s\eqc o$}}
      \rput(3.7,1){\rnode{jcm}{$m\eqc j$}}
      \rput(6.2,1){\rnode{csx}{$c\eqs X_\sortmale$}}
      \rput(3.5,2){\rnode{top}{$s(j,c)\eqc o(m,X_\sortmale)$}}
      \ncline{-}{jcm-abd}{jcm}
      \ncline{-}{csx-abd}{csx}
      \ncline{-}{sco-abd}{sco}
      \ncline{-}{jcm}{top}
      \ncline{-}{csx}{top}
      \ncline{-}{sco}{top}
    \end{pspicture}}
    \caption{Jon supported Clinton, but Mary opposed him.}
    \label{fig:clinton-der}
  \end{center}
\end{figure}

Given the above analysis, the relations $support$ and $oppose$ are
c-parallel, since they have both a common sort ($\sortsocial$) and a 
distinguishing sort ($\sortfriendly$). Further, in 
\begin{quote}
  {\it Jon supported Clinton, but Mary opposed him.}
\end{quote}
parallelism theory should predict that the full first utterance
{\it Jon supported Clinton} is c-parallel to the second namely, {\it
  Mary opposed him}. However, the sort $t$ does not have subsorts that
license this. Rather than dividing $t$ into cognitively unplausible
sorts, we propose an abductive equality calculus that generates all
possible explanations, why a pair of formulae could be c-parallel, based
on the respective sort hierarchies. The calculus manipulates two
equalities $\eqs$ for similarity and $\eqc$ for c-parallelism. The
inference rules given in figure~\ref{fig:parallel-calc}, give the
derivation from figure~\ref{fig:clinton-der} that explains the
parallelism in terms of assumed contrastivity and similarity of the
components. We have put the justifications of the abducibles in boxes.
Note that this calculus gives us the
explanation that {\it Jon supported Clinton} is c-parallel to {\it Mary
opposed him}, since {\it Jon} is c-parallel to {\it Mary} and {\it
support} is c-parallel to {\it oppose} and finally, we can make {\it
Clinton} and {\it him} similar by binding {\it him} to {\it Clinton}.

Of course, there is a similar derivation that makes {\it Mary} and
{\it Jon} similar and finally one that makes {\it support} and {\it
oppose} similar but {\it Mary} and {\it Jon} c-parallel. Thus we have
the problem to decide which of the different sets of abducibles is
the most plausible.

For this it is necessary to give a measure function for sets of abducibles.  For instance the three pairs
\[Jon\eqc Peter \qquad Jon\eqc Spot \qquad Jon\eqc\pi\]
are obviously ordered by increasing plausibility. We observe that this
plausibility coincides with the distance (the length of the connecting
path) from the least sorts of the objects to common sort. Therefore,
our approach is to derive plausibility values for abducibles from the
justifications of abducibles by calculating distances in the sort
hierarchies.

\section{HOU with Parallelism}
\label{sec:hou-par}

In recent approaches to ellipsis \cite{DalShiPer:eahou91} and
deaccenting \cite{GaKoLe:cahou96,Gardent:phouad97}, both parallelism
and higher-order unification are central to the semantic analysis. For
instance DSP analyze a VP-ellipsis such as
\begin{quote}
  {\it Jon likes golf, and Mary does too.}
\end{quote}
as being represented by $l(j,g)\land R(m)$ where $R$ represents the
ellipsis {\it does}, whose semantic value is then determined by solving the
equation $R(j)=l(j,g)$. The motivation for having $j$ occurr in the
left-hand side of the equation is that $j$ represents a c-parallel
element. This is where parallelism and the assumption of parallelism
structuration come in.  On the other hand, Higher-Order Unification is
also essential in that it is used to
solve the equation and furthermore, it is shown to be a crucial
ingredient in attaining wide empirical coverage (in particular, it is
shown to successfully account for the interaction of ellipsis with
quantification, anaphora and parallelism).

However, it is also known that a pure form of HOU is too powerful for
natural language and that a more restricted version of it namely,
Higher-Order Coloured Unification (HOCU) is more adequate in that it
helps prevent over-generation i.e. the prediction of linguistically
invalid readings \cite{GaKo:hocuanls96}. To see this, consider again
the example just discussed. Given the stipulated equation, HOU yields
two values for $R$ namely, $\lambda X. l(X,g)$ and $\lambda X.l(j,g)$,
of which only the first value is linguistically valid.  To remedy
this, DSP postulate a {\bf Primary Occurrence Restriction} (POR): the
term occurrence representing the element which is parallel to the
subject of the elliptical utterance, is a primary occurrence and any
solution containing a primary occurrence is discarded as
linguistically invalid. For instance, $j$ is a primary occurrence in
the equation $l(j,g) =R(j)$, so the solution $R=\lambda X.l(j,g)$
is invalid. \cite{GaKo:hocuanls96} show that DSP's POR can be expressed
within HOCU because it uses a variant of the simply typed
$\lambda$-calculus where symbol occurrences can be annotated with
so-called {\em colours} and substitutions must obey the following
constraint:
\begin{quote}
  For any colour constant \colc\ and any \colc --coloured variable
  $V_\colc$, a well--formed coloured substitution must assign to
  $V_\colc$ a \colc --monochrome term i.e., a term whose symbols are
 $\colc$--coloured.
\end{quote}
In this setting the POR can be expressed by coloring the primary
occurrence $j$ with a colour $\colpe$ but $R$ with a colour $\colnpe$.
Due to the constraint above, this in effect, enforces the POR.

More generally, \cite{GaKo:hocuanls96} argue that HOCU rather than
HOU, should be used for semantic construction as it allows a natural
modelling of the interface between semantic construction and other
linguistic modules. In what follows, we therefore assume HOCU as the
basic formalism and show how it can be combined with the abductive
calculus for parallelism, thereby providing an integrated framework
in which to handle parallelism, ellipsis and their interaction.

\subsection{Abductive Reconstruction of Parallelism}
\label{sec:abduc-par}

As just mentioned, we need a basic inference procedure that is a
mixture of higher-order colored unification and the sorted parallelism
calculus introduced above. The problem at hand is to make colored
sorted formulae similar or c-parallel.  For an algorithm ARP we build
up on a sorted version of HOCU (which can be obtained by a
straightforward combination of color techniques
from~\cite{HuKo:acvotlc95} with sorted methods
from~\cite{Kohlhase:amosho94})  but instead of simply having
equations for sorted $\beta\eta$-equality, we also add the
equations for c-parallelism and similarity to the unification
problem as special equations $\eqc$ and $\eqs$.

The ARP calculates sufficient conditions for a given set of input equations by
transforming systems of equations to a normal form from which these can be read off. 

Decomposition rules consist in those from
figure~\ref{fig:parallel-calc} with the
difference that   the rule for abstractions transforms equations of the form
$\lambda x.A=^t\lambda y.B$ to $[c/x]A=^t[c/y]B$, and $\lambda x.A=^tB$ to
$[c/x]A=^tBc$ where $c$ is a new constant, which may not appear in any
solution. Furthermore, there is a rule for colored constants that decomposes an
equation $c_{\cola}\eqs c_{\colb}$ into the color equation $\cola=\colb$.

The variable elimination process for colour variables is very simple, it allows to
transform a set ${\cal E}\cup\{\colavar=\cold\}$ of equations to
$[\cold/\colavar]{\cal E}\cup\{\colavar=\cold\}$, making the equation
$\{\colavar=\cold\}$ solved in the result. For the formula case, elimination is not
that simple, since we have to ensure that $|\sigma(x_\colavar)|=|\sigma(x_\colbvar)|$
to obtain a $\Col$-substitution $\sigma$. Thus we cannot simply transform a set
${\cal E}\cup\{x_\cold=M\}$ into $[M/x_\cold]{\cal E}\cup\{x_\cold=M\}$, since
this would (incorrectly) solve the equations $\{x_\colc=f_\colc,x_\cold=g_\cold\}$.
The correct variable elimination rule transforms ${\cal E}\cup\{x_\cold=M\}$ into
$\sigma({\cal E})\cup\{x_\cold=M,x_{\colc_1}=M^1,\ldots,x_{\colc_n}=M^n\}$, where
$\colc_i$ are all colours of the variable $x$ occurring in $M$ and $\cal E$, the
$M^i$ are appropriately coloured variants (same colour erasure) of $M$, and $\sigma$
is the $\Col$-substitution that eliminates all occurrences of $x$ from $\cal E$.

Due to the presence of function variables, systematic application of these rules can
terminate with equations of the form
$x_\colc(s^1,\ldots,s^n)=h_\cold(t^1,\ldots,t^m)$.  Such equations can neither be
further decomposed, since this would loose unifiers (if $G$ and $F$ are variables,
then $Ga=Fb$ as a solution $\lambda x.c$ for $F$ and $G$, but $\{F=G,a=b\}$ is
unsolvable), nor can the right hand side be substituted for $x$ as in a variable
elimination rule, since the sorts would clash. The sorted, colored variant of Huet's
classical solution to this problem is to instantiate $x_{\colc}$ with a
$\colc$-monochrome formula that has the right sort $\overline{\sortb_n}\ar\sorta$
(that of $x_{\colc}$) and the right head $h_\cold$ (which we assume to have sort
$\overline{\gamma_m}\ar\sorta$). These so-called {\bf general bindings} have the
following form:
\[{\cal G}^h_\cold=\lambda z^{\sorta_1}\ldots z^{\sorta_n}.h_\cold(H_{\cole_1}^1(\overline{z}),\ldots,H_{\cole_m}^m(\overline{z}))\]
where the $H^i$ are new variables of sort $\overline{\sortb_n}\ar\gamma_i$ and the
$\cole_i$ are either distinct colour variables (if $\colc\in\Covars$) or
$\cole_i=\cold=\colc$ (if $\colc\in\Colours$). If $h$ is one of the bound variables
$z^{\sorta_i}$, then ${\cal G}^h_\cold$ is called an {\bf imitation binding}, and
else, ($h$ is a constant or a free variable), a {\bf projection binding}.

The general rule for flex/rigid equations transforms
$\{x_\colc(s^1,\ldots,s^n)=h_\cold(t^1,\ldots,t^m)\}$ into
$\{x_\colc(s^1,\ldots,s^n)=h_\cold(t^1,\ldots,t^m), x_\colc={\cal G}_\colc^h\}$,
which in essence only fixes a particular binding for the head variable $x_\colc$. It
turns out (for details and proofs see~\cite{HuKo:acvotlc95}) that these general
bindings suffice to solve all flex/rigid situations, possibly at the cost of creating
new flex/rigid situations after elimination of the variable $x_\colc$ and
decomposition of the changed equations (the elimination of $x$ changes
$x_\colc(s^1,\ldots,s^n)$ to ${\cal G}^h_\colc(s^1,\ldots,s^n)$ which has head $h$).
This solution for pure equations has to be adapted to the more general similarity and
contrastivity relations $\eqs$ and $\eqc$, where we have to provide further imitation
rules. In particular, for an equation $X^\sorta_\cola\ov\bU\eqc h\ov\bV$ we have to
allow imitation bindings $\cG^k_{\sorta,\cola}$ for $X^\sorta_\cola$ for any constant
$k$ that is contrastive to $h$ and analogously for $\eqs$.

\subsection{Gapping and ARP}
\label{sec:gapping}

We now illustrate the workings of ARP by the following 
example
\begin{quote}
  {\it Jon likes golf, and Mary too.}
\end{quote}
where the second clause is a gapping clause in that both the verb and
a complement are missing. This example clearly illustrates the
interaction of parallelism with semantic interpretation: if the
parallel elements are {\it Jon} and {\it Mary}, the interpretation of
the gapping clause is {\it Mary likes golf}, but if conversely {\it
Mary} is parallel to {\it golf}, then the resulting interpretation is
{\it Jon likes Mary}. Although the first reading is clearly the
default, the second can also be obtained -- in a joke context for
instance. In what follows, we show that ARP predicts both the
ambiguity and the difference in acceptability between the two possible
readings. Additionally, we show that DSP's a-priori labelling of
occurrences as primary or not primary can now be reduced to a more
plausible constraint namely, the constraint that {\it Mary} is a
parallel element which has exactly one parallel counterpart in the
source (or antecedent) clause.

The analysis is as follows. First, we
follow DSP and assign the above example the representation
\[l(j,g)\land R(m)\] where $R$ stands for the missing semantics.
However, in contrast to DSP, we do not presuppose any knowledge about
parallelism in the source utterance and determine the meaning of $R$
from the equation
\[l(j_\colavar,g_{\neg\colavar})\eqc
R_\colnpe^{\sortwoman\ar\typebool}(m_\colpe)\] which only says that
the propositions expressed by {\it Jon likes Mary} and {\it golf too}
stand in a c-parallel relation\footnote{By contrast, an extension of
  DSP's analysis to gapping would posit the equations $l(j,g) = R(j)$
  and $l(j,g) = R(g)$ thereby postulating 
  both the parallel elements, and the ambiguity of the gapping
  clause.}. The rationale for the colors in this equation is that {\it
  Mary} must be a parallel element in the target utterance. For {\it
  Jon} and {\it golf} in the source utterance, we do not know yet
which of them will be a parallel element, but it can be at most one of
them, which we code by giving them unspecified but contradictory
colors\footnote{Clearly, this coding is not general enough for the
  general case, where there are more than one parallel elements in the
  target utterance, we leave a general treatment to further work.}.
Finally, $R$ gets the color $\colnpe$, since it may not be
instantiated with formulae that contain primary material (POR).

Since the elided material in gapping constructions and VPE may only copy material from the source
utterance (and may not introduce new material) we add the constraint to ARP that $\eqc$ and $\eqs$
imitations may only be applied to equations, where the head is $\colpe$-colored. We call this the
{\bf copying constraint} for gapping and VPE. It ensures that whenever two elements are similar but not identical, then they must be primary, since they are parallel.

Let us now go through the ARP computation to see that our analysis obtains exactly the desired
readings and to gain an insight of the mechanisms employed therein. 

The initial equation is a flex/rigid pair, where only the strict imitation\footnote{Note the copying
  constraint is at work here.} rule is applicable (there is no projection binding of sort
$\sortwoman\ar\typebool$). So, we obtain the binding $\lam Z.l(H_\colnpe Z)(K_\colnpe Z)$, where $H$
and $K$ are new variables of sort $\sortwoman\ar\sorthuman$. Eliminating this equation yields the
equation
\[l(j_\colavar,g_{\neg\colavar})\eqc l(H_\colnpe m_\colpe)(K_\colnpe m_\colpe)\]
which can be decomposed to the equations 
\[H_\colnpe m_\colpe\eqc j_\colavar  \hspace{3cm} K_\colnpe m_\colpe\eqc g_{\neg\colavar}\]
For the variable $H_\colnpe$ in the first equation both the imitation
binding $\lam Z.j_\colnpe$ and the projection binding $\lam Z.Z$ are
possible.

In the first case, we have the equation $j_\colavar\eqc j_\colnpe$,
which entails that $\colavar=\colnpe$ leaving us with the second
equation (we can eliminate double negations on colors)
\[K_\colnpe m_\colpe\eqc g_\colpe\]
Again we have the possibility of imitate or project. Since the
imitation binding $\lam Z.g_\colpe$ for $K_\colpe$ leads to a color
clash in $g_\colpe\eqc g_\colnpe$, only the projection binding $\lam
Z.Z$ yields a solution, since the resulting equation $m_\colpe\eqc
g_\colpe$ is valid, since {\it golf} and {\it mary} share the sort
$\sortreal$.

If, on the other hand, we choose the projection binding for
$H_\colpe$, then variable elimination yields the equation
$j_\colavar\eqc m_\colpe$, which is valid, since {\it Jon} and {\it
Mary} share the sort $\sorthuman$ and which entails that
$\colavar=\colpe$ leaving us with the second equation
\[K_\colnpe m_\colpe\eqc g_\colnpe\]
Again we have the possibility of imitate or project. This time, the imitation binding $\lam
Z.g_\colnpe$ for $K_\colnpe$ leads to the trivialy valid equation $g_\colnpe\eqc g_\colnpe$, while the
projection binding $\lam Z.Z$ yields the equation $m_\colpe\eqc g_\colnpe$, which must be unsolvable,
since the colors clash.

If we collect the bindings, we arrive at the two solutions $\lam
Z.l(Z,g)$ and $\lam Z.l(j,Z)$, which correspond to the readings {\it
  Mary likes golf} and {\it Jon likes Mary}. Note that since the
similarity of {\it Jon} and {\it Mary} is stronger than that between {\it
  Mary} and {\it golf}, the first reading is preferred, while the
second reading may only be obtained in the context of a joke.
Note also that the use of colours (i.e. the constraint that {\it Mary}
has exactly one parallel counterpart in the source) correctly rules
out the mathematically valid solution $\lam Z.l(Z,Z)$ where {\it Mary}
would be analysed as contrasting with both {\it Jon} and {\it golf}.

We now briefly compare our approach with~\cite{GroBreManMoe:puagidg94}
where an account of VP-ellipsis is given which uses first-order
default unification on feature-structure semantic representations and
claims to predict the parallel elements.  The underlying idea is that
parallel elements have identical thematic roles. Although the approach
works well for VP-ellipsis, it is only because in the VPE case, the
parallel element of the elliptical clause is known to fill the
agentive role. However in gapping cases, part of the resolution
problem is to determine the thematic role of the constituents
appearing in the elliptical clause.  For instance, in {\it Jon likes
  golf, and Mary too}, the thematic role of {\it Mary} is either
agent or patient. Unfortunately, the only obvious way to incorporate
this into a default unification approach would be to augment it with
variables over features -- an essentially higher-order construct.

\subsection{Controlling ARP}
\label{sec:control}

Clearly, a naive implementation of the ARP calculus as sketched above
will be intractable, since the set of abducibles is much too large.
However, most abducibles are very implausible and should not be
considered at all. As in all implementations of abductive processes,
the search for abducibles has to controlled, which in turn calls for a
quality measure of abduced equations. A standard (but not very
imaginative\footnote{Clearly, a more sophisticated measure would
  include concepts like the specificity of the solution.}) measure
would be the conceptual distance of the sorts justifying the equation,
(i.e.  the number of subsorts crossed to reach the common and
discerning sorts). In our example, the rating of $m\eqc g$ is 6, while
that of $m\eqc j$ is 2, justifying the claim that the reading {\it
  Mary likes golf.} is more plausible than {\it Jon likes Mary.}
Since all other readings are either ruled out by the colors or are
even more implausible, e.g. an $A^{*}$ implementation of ARP will only
derive these, iff given an appropriate threshold. Since the aim of
this paper is to establish the principles of parallelism
reconstruction, we will not pursue this here.

\section{Conclusion}
\label{sec:concl}

We have given a sketch of how to develop a computational framework for
calculating parallelism in discourse. This approach is based on the
HOCU variant of DSP's HOU account of ellipsis, but unlike that
approach does not presuppose knowledge about the parallel elements.
Instead, it computes them in the analysis.

Parallelism can be seen as affecting the interpretation of the second
of two parallel utterances in mainly two ways: it can either
constrain an anaphor to resolve to its source parallel counterpart
(this is the case for instance, in the gapping example discussed
above); or it can add to its truth conditional content. For instance,
in
\begin{quote}
{\it Jon campaigned hard for Clinton in 1992. Young aspiring politicians
often support their party's presidential candidate}
\end{quote}
parallelism enforces a reading such that {\it Jon} is understood to be
a {\it young aspiring politician} and {\it Clinton} is understood to
be Jon's {\it party's presidential candidate}. 

In future work, we plan to investigate these two aspects in more
details. As for the interaction of parallelism with binding, one
important question is whether our proposal preserves DSP's insights on
the interaction of parallelism with ellipsis, anaphora and
quantification. On the other hand, to account for the incrementing
effect of parallelism on semantic interpretation, the proposal will
have to cover the discourse relations of exemplification and
generalisation. Note however that the proposed interleaving between
HOU and abductive calculus gives us a handle on that problem:
mismatches between semantic structures can be handled by having the
calculus be extended to abstract away irrelevant structural
differences (this would account for instance for the fact that in our
example, a temporal modifier occurs in the source but not in the
target) whereas sorted HOU can be used to infer information from the most
specific common sort (in this case, the sort of {\it young aspiring
  politicians}).


\end{document}